\documentclass[letterpaper, 10 pt, conference]{ieeeconf}
\IEEEoverridecommandlockouts

\newcommand\copyrighttext{%
	\footnotesize This work has been submitted to the IEEE for possible publication. Copyright may be transferred without notice, after which this version may no longer be accessible.
}
\newcommand\copyrightnotice{%
	\begin{tikzpicture}[remember picture,overlay]
	\node[anchor=south,yshift=10pt, xshift=10pt] at (current page.south) {\fbox{\parbox{\dimexpr\textwidth-\fboxsep-\fboxrule\relax}{\copyrighttext}}};
	\end{tikzpicture}%
}

\usepackage{cite}
\usepackage{amsmath,amssymb,amsfonts}
\usepackage{algorithm}
\usepackage{algpseudocode}
\usepackage{graphicx}
\usepackage{comment}
\usepackage{url}
\usepackage{hyperref}
\usepackage{textcomp}
\usepackage{xcolor}
\usepackage[nolist]{acronym}
\usepackage{tikz}
\usepackage{siunitx}
\usetikzlibrary{shapes.geometric, arrows, positioning}
\usepackage{booktabs} 
\usepackage{multirow} 
\usepackage[caption=false,font=scriptsize]{subfig} 

\definecolor{Blue}{HTML}{0065bd}
\definecolor{Bluelight}{HTML}{2AD5F3}
\definecolor{FerrariRed}{HTML}{da1919}

\graphicspath{{./figs/}}

\tikzstyle{block} = [rectangle, rounded corners, minimum width=3cm, minimum height=1cm,text centered, draw=black, fill=blue!20]
\tikzstyle{input} = [ellipse, minimum width=2cm, minimum height=1cm, text centered, draw=black, fill=green!20]
\tikzstyle{output} = [ellipse, minimum width=2cm, minimum height=1cm, text centered, draw=black, fill=red!20]
\tikzstyle{arrow} = [thick,->,>=stealth]

\def\BibTeX{{\rm B\kern-.05em{\sc i\kern-.025em b}\kern-.08em
    T\kern-.1667em\lower.7ex\hbox{E}\kern-.125emX}}
\begin{document}

\title{Reinforcement Learning-based Dynamic Adaptation\\for Sampling-Based Motion Planning in Agile Autonomous Driving}

\author{Alexander Langmann, Yevhenii Tokarev, Mattia Piccinini, Korbinian Moller and Johannes Betz 
\thanks{A. Langmann, Y. Tokarev, M. Piccinini, K.Moller and J. Betz are with the Professorship of Autonomous Vehicle Systems, TUM School of Engineering and Design, Technical University of Munich, 85748 Garching, Germany; Munich Institute of Robotics and Machine Intelligence (MIRMI)}
}

\maketitle
\copyrightnotice
\begin{abstract}
Sampling-based trajectory planners are widely used for agile autonomous driving due to their ability to generate fast, smooth, and kinodynamically feasible trajectories. However, their behavior is often governed by a cost function with manually tuned, static weights, which forces a tactical compromise that is suboptimal across the wide range of scenarios encountered in a race. To address this shortcoming, we propose using a Reinforcement Learning (RL) agent as a high-level behavioral selector that dynamically switches the cost function parameters of an analytical, low-level trajectory planner during runtime. 
We show the effectiveness of our approach in simulation in an autonomous racing environment where our RL-based planner achieved 0\% collision rate while reducing overtaking time by up to 60\% compared to state-of-the-art static planners. Our new agent now dynamically switches between aggressive and conservative behaviors, enabling interactive maneuvers unattainable with static configurations. These results demonstrate that integrating reinforcement learning as a high-level selector resolves the inherent trade-off between safety and competitiveness in autonomous racing planners. The proposed methodology offers a pathway toward adaptive yet interpretable motion planning for broader autonomous driving applications.

\end{abstract}


\section{Introduction}
\subsection{Motivation}
The field of autonomous racing pushes motion planning algorithms to their absolute limits, requiring them to operate safely and adaptively under challenging conditions \cite{betz2022Autonomous}. A popular and effective solution is the use of sampling-based planners in a Frenet frame, which generates smooth, feasible trajectories for high-speed maneuvers \cite{werling2010Optimal}, \cite{ogretmen2024SamplingBasedb}.

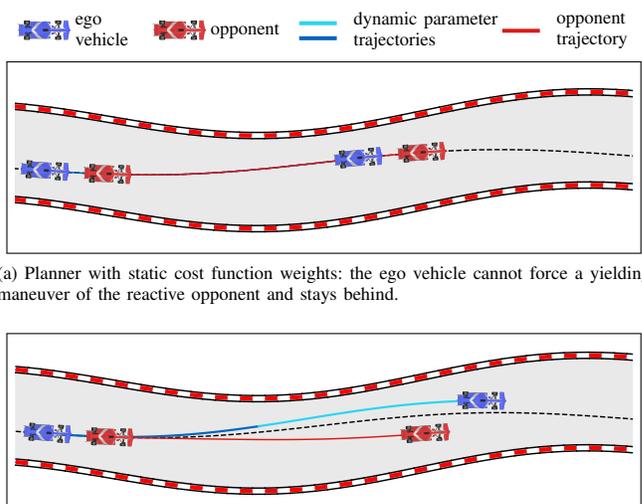
\begin{figure}[!t]
    \centering
    \subfloat[][Planner with static cost function weights: the ego vehicle cannot force a yielding maneuver of the reactive opponent and stays behind.]{
        \label{fig:RL_intro_a}
        \begin{tikzpicture}[font=\scriptsize]
            
            \node[inner sep=0pt] at (0.5,0) {\includegraphics[height=2.5mm]{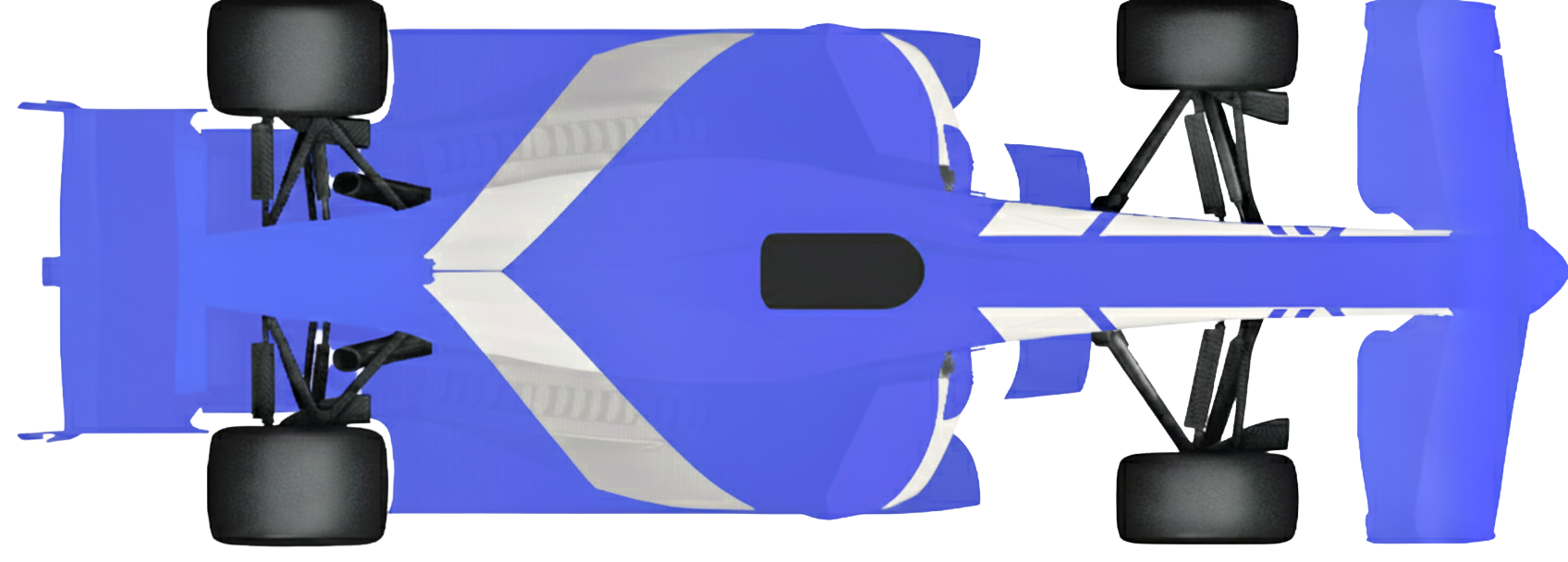}};
            \node[align=left, anchor=west] at (0.8,0) {ego \\ vehicle};
        
            \node[inner sep=0pt] at (2.3,0) {\includegraphics[height=2.5mm]{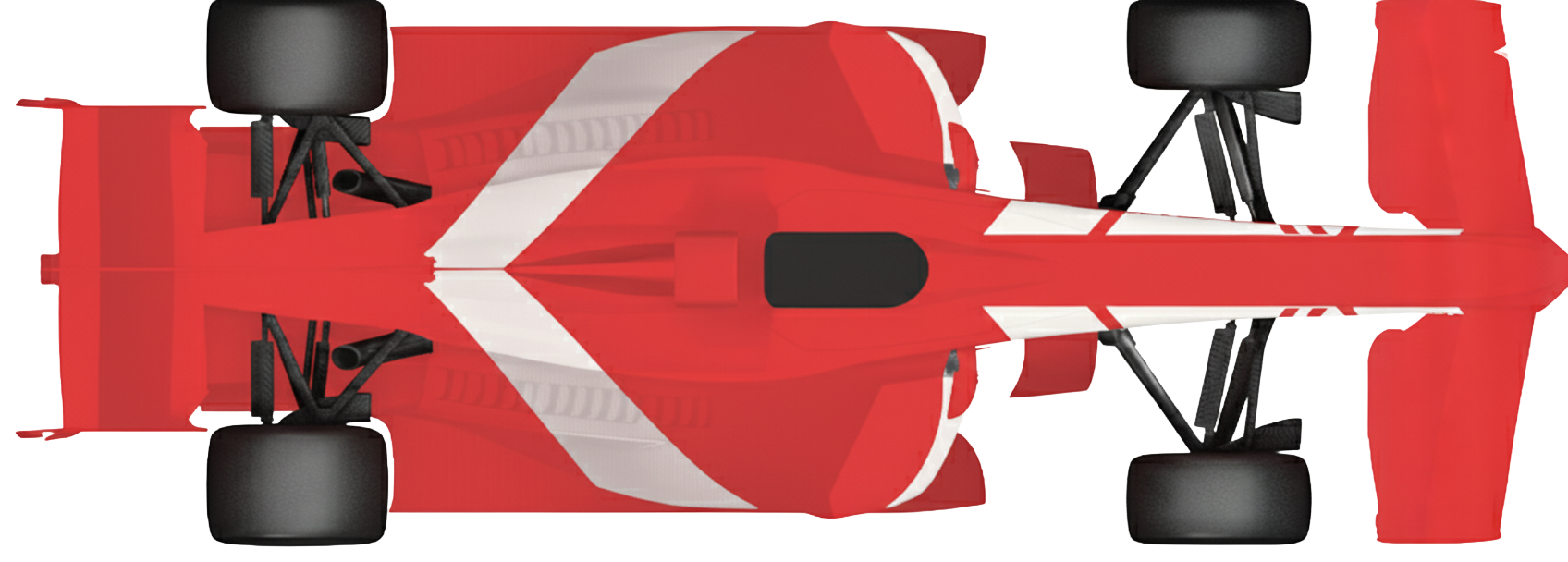}};
            \node[align=left, anchor=west] at (2.6,0) {opponent};
    
            \draw[thick, Bluelight, line width=1.5pt] (3.9,0.1) -- (4.4,0.1);
            \draw[thick, Blue, line width=1.5pt] (3.9,-0.1) -- (4.4,-0.1);
            \node[align=left, anchor=west] at (4.5, 0) {dynamic parameter \\ trajectories};
    
            \draw[thick, FerrariRed, line width=1.5pt] (6.6,0.0) -- (7.1,0.0);
            \node[align=left, anchor=west] at (7.2, 0) {opponent \\trajectory};

            \node[anchor=north west, inner sep=0pt] at (0,-0.4) {\fbox{\includegraphics[width=0.95\linewidth, trim={0cm 1.5cm 0cm 1.5cm},clip]{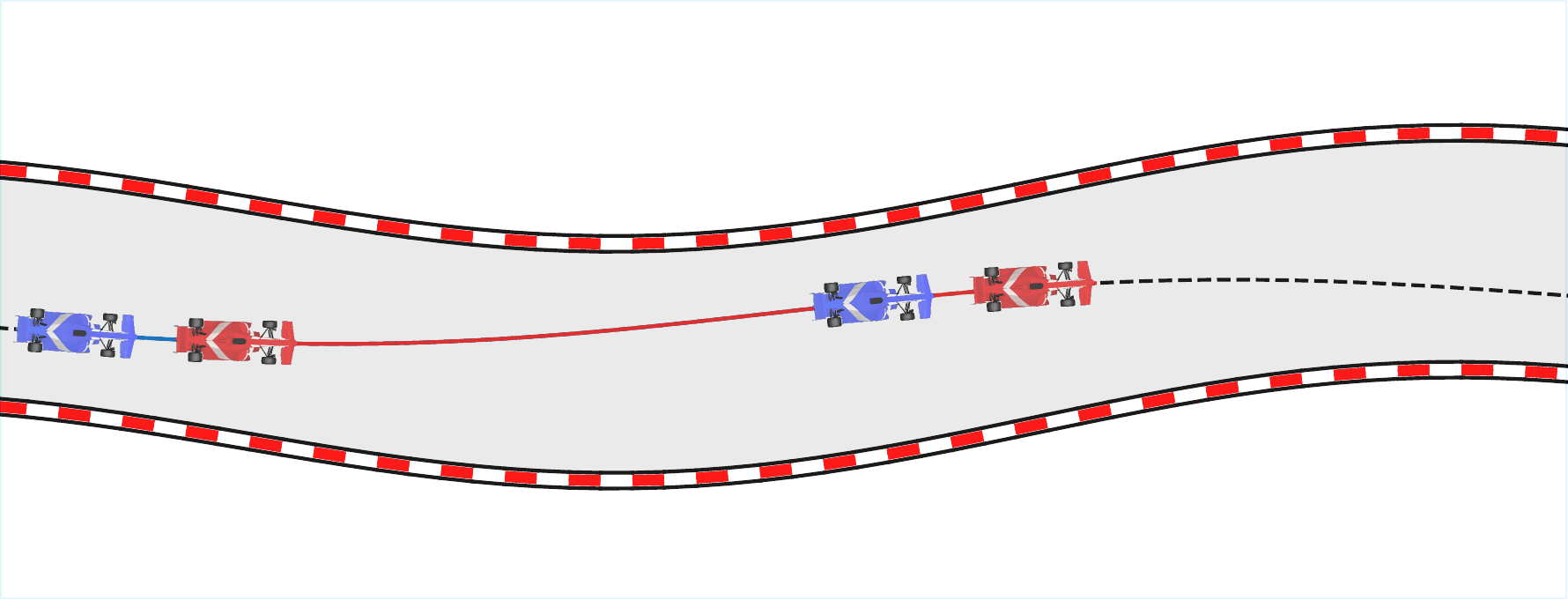}}};
        \end{tikzpicture}
    } \\
    \subfloat[][Planner with adaptive cost function weights: the ego vehicle anticipates the yielding maneuver of the reactive opponent when committing to an aggressive trajectory, which results in a successful overtake.]{
        \label{fig:RL_intro_b}
        \fbox{\includegraphics[width=0.95\linewidth, trim={0cm 1.5cm 0cm 1.5cm},clip]{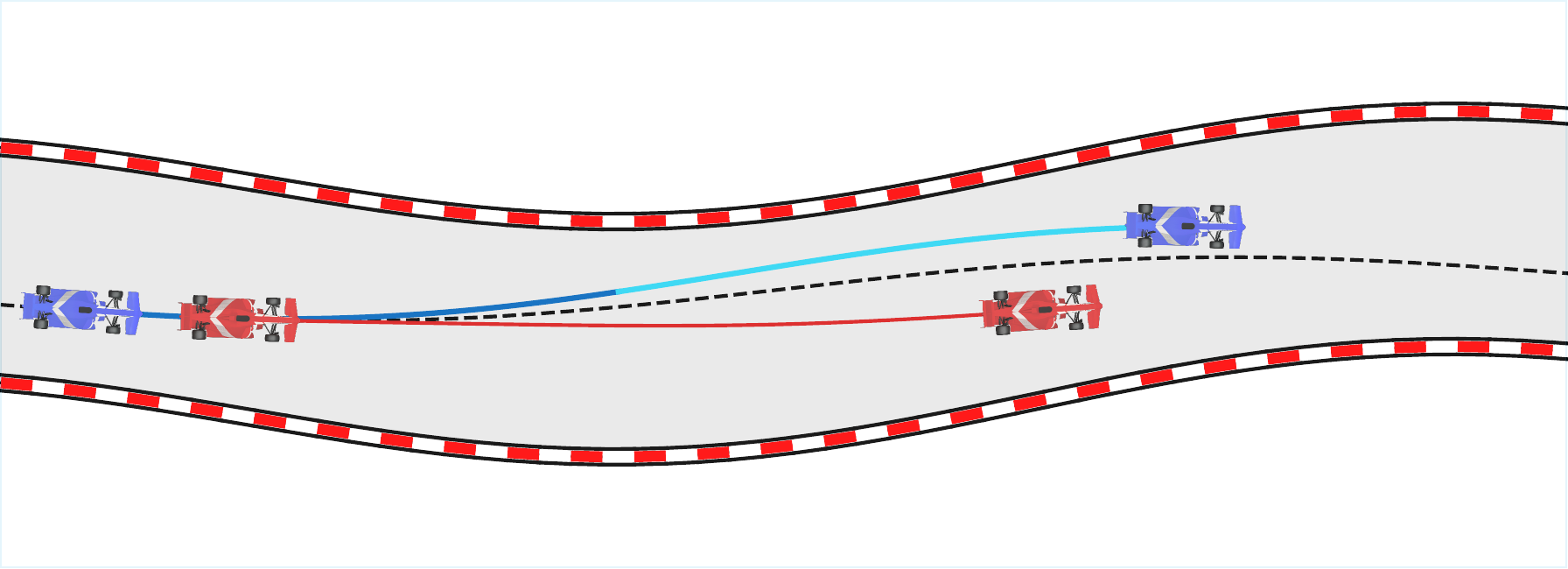}}
        }
    \caption{Comparison of our RL-based planner with dynamic weight adaptation against a static-weight planner.}
    \label{fig:RL_intro}
\end{figure}

The behavior of these planners is fundamentally dictated by a cost function that evaluates candidate trajectories based on metrics such as speed, deviation from the racing line, and proximity to opponents. The weights of this function are typically tuned manually and remain static throughout operation \cite{Sampling_TUning2021}. This presents a significant limitation: a single set of weights represents a fixed behavioral compromise. For instance, a configuration tuned for optimal single-vehicle performance may be dangerously aggressive in multi-vehicle environments.
Our goal is to enable the possibility of anticipating an opponent vehicle's behavior and then forcing the vehicle into a dedicated yielding maneuver while overtaking (Figure \ref{fig:RL_intro}). Therefore, our ego vehicle must plan a trajectory that temporarily lowers its own collision cost parameter -- a risk too great to encode in a single, static configuration.  

To address this rigidity, we propose a trajectory planner where a high-level \ac{RL} agent acts as a behavior selector for the low-level sampling-based planner. From a predefined library of cost function parameter sets, the agent chooses the most appropriate set at each time step. Each set in this library favors a distinct driving behavior suitable for different scenarios. Because the planner can only operate within these predefined modes, its worst-case behavior remains predictable. This approach maintains the stability and robustness of the analytical planner while enabling situational adaptability that cannot be achieved by static parameters.

\begin{figure*}[!t]
   \centering
    \includegraphics[width=1.0\textwidth]{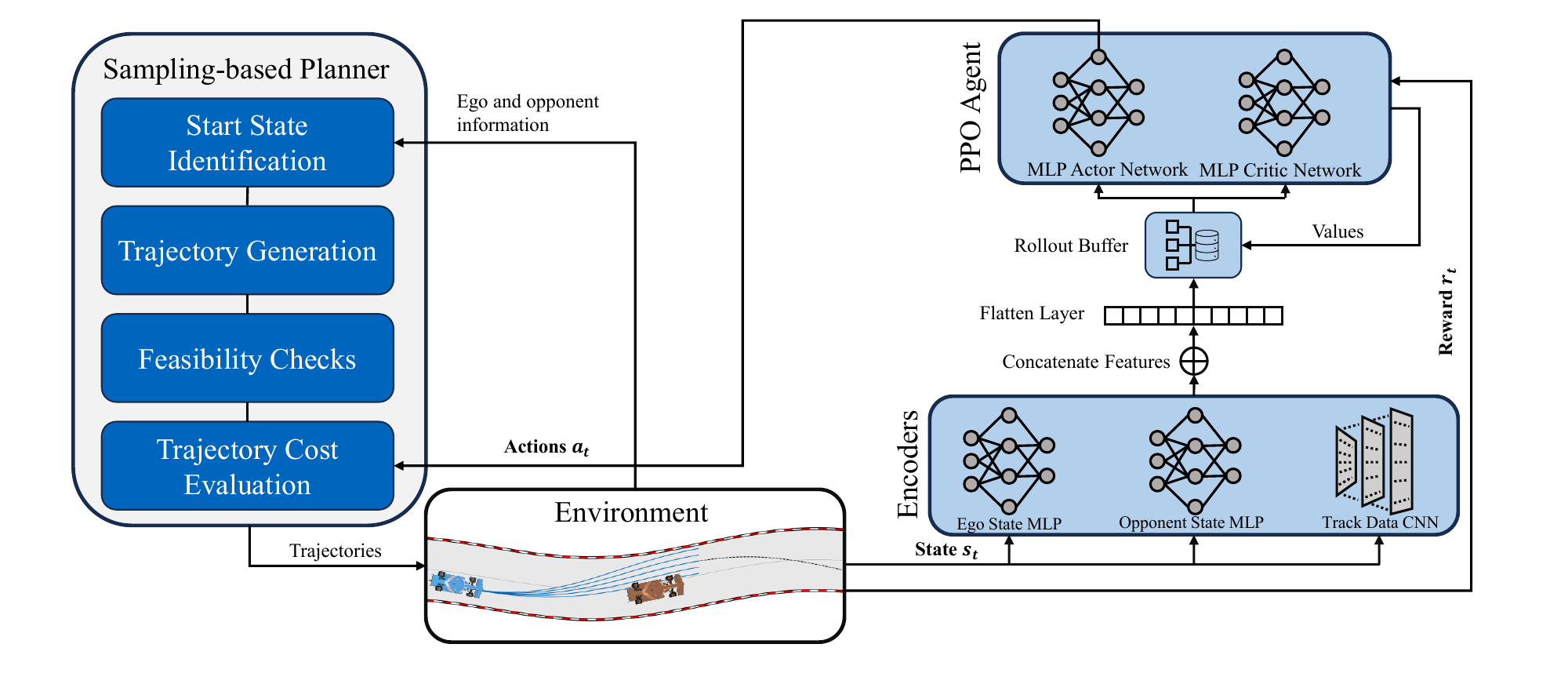}
    \caption{Framework overview: Our RL agent observes the states of the ego vehicle and its closest opponent,  along with track-dependent features, which are processed by a PPO-based actor–critic model.
    The RL actions $a_{t}$ adapt the cost function parameters of a sampling-based planner to boost the performance in dynamic overtaking maneuvers.
    }\label{fig:method_overview}
\end{figure*}

\subsection{Related Work}
Trajectory planning for high-performance autonomous vehicles is predominantly handled by classical methods, including graph-based \cite{stahl2019Multilayer}, optimization-based \cite{vazquez2020OptimizationBased, Liniger2014, Reiter2021}, and sampling-based approaches \cite{Arslan2017, ogretmen2024SamplingBasedb}. Among these, sampling-based methods operating in a Frenet frame are widespread in racing applications due to their robustness and ability to generate kinodynamically feasible trajectories \cite{piazza2024MPTreea, ogretmen2024SamplingBasedb, raji2022Motiona, jung2023Autonomous}. The core limitation of these planners stems from their inherent "predict-then-plan" logic: Opponents are treated as dynamic obstacles on a fixed future path, not as strategic agents whose intentions and actions might change. This renders the planner's behavior purely reactive. This reactive nature is solidified by a static cost function, whose manually-tuned weights must encode a single, fixed assumption about opponent behavior for all situations. This forces a suboptimal compromise that fundamentally limits the vehicle's adaptability and prevents it from executing truly interactive maneuvers. One possibility for overcoming this limitation could be interaction-based game theory \cite{Wang2019, wang2020, Liniger2020, rowold2024open}, but existing formulations are often too slow, rely on pre-modeled opponent behavior, and lack the interactivity required for real-time racing scenarios.

\ac{RL} approaches have recently been deployed to improve adaptability and interaction capabilities in motion planning for autonomous racing \cite{evans2023Comparing}. In racing simulation environments, these approaches have demonstrated super-human performance by mapping sensor inputs directly to low-level control commands \cite{wurman2022Outracinga, fuchs2021SuperHuman}. These methods can learn highly complex and interactive maneuvers that are difficult to hand-craft, as shown in competitive racing games \cite{song2021Autonomousa} and dense traffic scenarios \cite{Masayoshi2019}. However, this performance often comes at the cost of safety guarantees. As these policies operate as "black boxes", it is difficult to verify their behavior or predict their actions in novel situations. Further, they often require millions of simulation steps to converge, and while transfer from simulation to real-world application has been successfully shown in drone racing \cite{kaufmann2023Championlevel}, this transfer poses an unsolved problem for real-world autonomous car racing.

To combine the strengths of classical planners and \ac{RL}-based approaches, several hybrid architectures have been proposed both for road traffic and racing environments. 
Here, the \ac{RL}-agent is employed to execute a subtask of the motion planner. For road traffic scenarios, this has been applied to select cost function weights \cite{trauth2024Reinforcement} or trajectory parameters such as target velocity and lateral position \cite{mirchevska2023Optimizing} online. In autonomous racing, \ac{RL}-based end condition selection has been used to perform interactive overtakes on a blocking opponent on a straight track section \cite{ogretmen2024Trajectory}. \ac{RL}-assisted cost function parametrization has also been explored in other autonomous driving domains, such as motion control \cite{zarrouki2024Safe},  \cite{dwivedi2022Continuous}.
To our understanding, the state-of-the-art is limited by an interactive trajectory planning method that utilizes the adaptability reachable by using \ac{RL} without losing interpretability in a complex autonomous racing domain. 
In summary, while existing methods either lack adaptability or sacrifice interpretability, there remains a clear need for an interactive, safe, and dynamically adaptive trajectory planning approach for agile autonomous driving.

\subsection{Contributions}
From the research gap derived above, we identify the following three key contributions:
\begin{itemize}
    \item An RL-based trajectory planner that enables interactive maneuvers in autonomous racing. Our \ac{RL} agent learns to dynamically adapt cost function parameters, guiding the planner to select a suitable behavior according to the race scenario.
    \item A methodology for integrating learning-based components that ensures trajectory validity by design, preserving the safety guarantees of the deterministic planner.
    \item A demonstration in simulation that this dynamic behavioral adaptation significantly outperforms any single, static parameter configuration in challenging, interactive multi-vehicle racing scenarios on two different racetracks.
\end{itemize}

\section{Methodology}
Figure \ref{fig:method_overview} gives an overview of our framework, which consists of two layers: a low-level sampling-based trajectory planner that ensures safety and feasibility, and a high-level \ac{RL} agent that boosts the performance in dynamic overtaking maneuvers, by adapting the planner's cost function weights online. 

\subsection{Sampling-based Trajectory Planner}
\label{subsec:sb_planner}
We use a sampling-based trajectory planner that operates in a Frenet frame in a three-dimensional track representation \cite{ogretmen2024SamplingBasedb, langmann2025Online}. We denote $s$ as the longitudinal progress along the reference line, which is an offline-optimized raceline, and $n$ as the lateral deviation from this reference line at a given coordinate $s$. 
Each planning step consists of four stages (Figure \ref{fig:method_overview}): start state identification, trajectory generation, feasibility checks, and trajectory cost evaluation. 
From the identified start state, the planner generates a set of candidate trajectories by sampling terminal states in longitudinal velocity $\dot{s}_{\mathrm{end}}$ and lateral position $n_{\mathrm{end}}$ over a fixed time horizon $T$. Quartic and quintic polynomials are used to generate jerk-minimal longitudinal and lateral profiles, respectively, from the current planning start state to the sampled terminal states \cite{werling2010Optimal}. 
To be considered for execution, a generated trajectory must be kinodynamically feasible. This is ensured by validating that every point along the trajectory adheres to the following set of hard constraints:
\begin{equation}
\label{eq:feasibility_checks}
\begin{aligned}
    &\text{Curvature:} & \kappa &\leq \kappa_{\mathrm{max}},\\
    &\text{Speed limits:} & v &\leq v_{\mathrm{max}},\\
    &\text{Track limits:} & n_{\mathrm{min}} \leq n &\leq n_{\mathrm{max}},\\
    &\text{Engine limits:} & \tilde{a}_{\mathrm{x}} &\leq \tilde{a}_{\mathrm{x, eng}},\\
    &\text{Acceleration:} & \left(\frac{\tilde{a}_{\mathrm{x}}}{\tilde{a}_{\mathrm{x, max}}}\right)^{p} + \left(\frac{\tilde{a}_{\mathrm{y}}}{\tilde{a}_{\mathrm{y, max}}}\right)^{p} &\leq 1.
\end{aligned}
\end{equation}
where $\tilde{a}_x$ and $\tilde{a}_y$ are the apparent longitudinal and lateral accelerations, following the definition provided in \cite{lovato_three-dimensional_2022}, and $p$ determines the gg-diagram shape used for limit evaluation. Note that the time dependency for all variables has been omitted for brevity. Any trajectory that violates even one of these conditions is deemed infeasible and is immediately discarded.

The remaining feasible trajectories are evaluated using a cost function to find the optimal candidate:
\begin{equation}
    C = \int^{T}_{0} (w_{\mathrm{rl}}c_{\mathrm{rl}} + w_{\mathrm{v}}c_{\mathrm{v}} + w_{\mathrm{a}}c_{\mathrm{a}} + w_{\mathrm{pr}}c_{\mathrm{pr}} + w_{\mathrm{c}}c_{\mathrm{c}}) \ \text{d}t.
    \label{eq:cost_fun}
\end{equation}
The terms penalize deviations from the optimal raceline ($c_{\mathrm{rl}}$) and target velocity ($c_{\mathrm{v}}$), violations of the acceleration limits ($c_{\mathrm{a}}$), close proximity to opponents ($c_{\mathrm{pr}}$), and high collision risk ($c_{\mathrm{c}}$). The planner's tactical behavior is determined by the weights $w$, which balance these competing objectives.

\subsection{Reinforcement Learning Framework}
\label{subsec:rl_procedure}
\begin{figure}[b]
    \centering
    \includegraphics[width=1.0\linewidth]{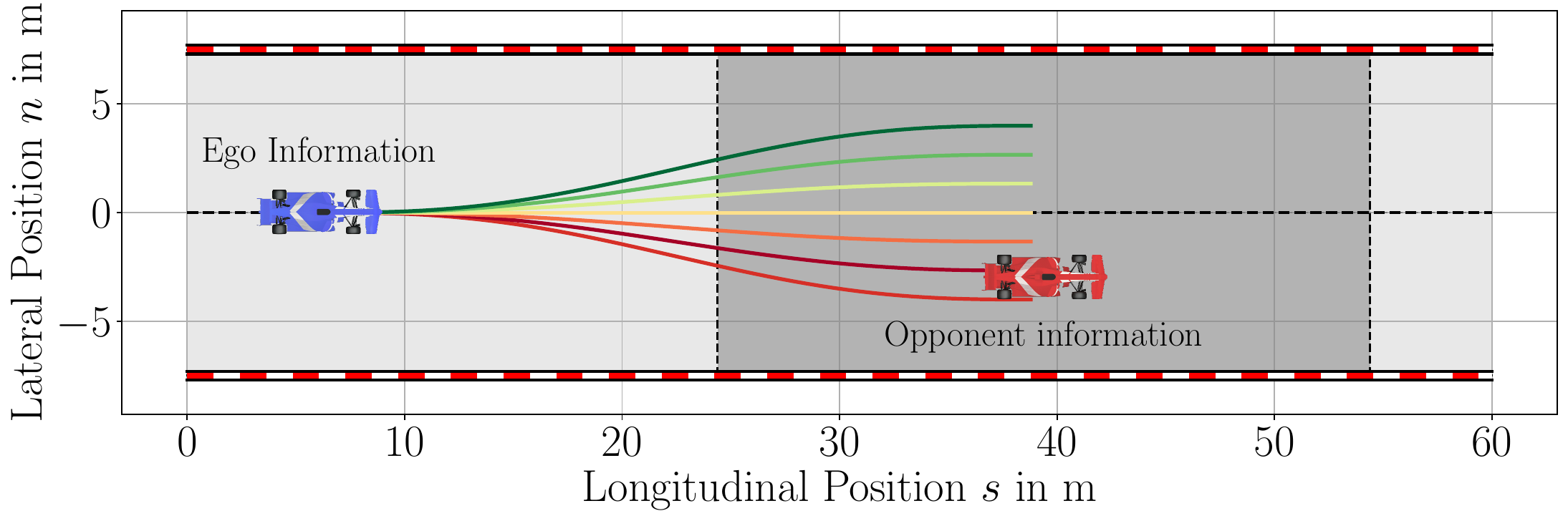}
    \caption{Representation of the interaction zone (dark grey area) that triggers the active terms in the RL reward function. When the ego vehicle enters this zone, the reward logic activates gap and collision penalties, and fixed raceline following is suppressed. Trajectories are color-coded by costs from green (cheap) to red (expensive). 
    }
    \label{fig:scene_design}
\end{figure}

We formulate the task of selecting the optimal cost function weights as an \ac{MDP}. The agent is trained over multiple \textit{episodes}, where an episode is defined as a complete scenario run (e.g., a full lap or a specific overtaking maneuver). Each episode consists of multiple \textit{iterations}, where each iteration corresponds to a single planning step of the low-level planner. Figure \ref{fig:scene_design} shows an overview of an exemplary scenario.

\textbf{Observation Space:} The observation space $\mathcal{O}$ provides the agent with a comprehensive view of the current environment, and is described in Table~\ref{tab:state_space}. It consists of the ego state, the state of the closest opponent, and the track representation (Figure \ref{fig:method_overview}), concatenated into a single vector. To ensure consistent scaling for the neural network, all variables within this vector are normalized in the range [-1, 1].
The ego state includes the vehicle's kinematic state, including position, velocity, and acceleration. The opponent's state contains the relative position and velocity of the nearest opponent vehicle within a defined detection range. Key features of the track representation are summarized in Table~\ref{tab:state_space}, and include the track boundaries and a summary of the upcoming reference line's 3D geometry $\textbf{G}(s)$, which encompasses features like banking, slope, yaw angle, and curvature.

\begin{table}[t]
\centering
\caption{Observation space of our RL agent.}
\label{tab:state_space}
\begin{tabular}{@{}llcl@{}}
\toprule
\textbf{Category}    & \textbf{Obs. Space} & \textbf{Unit} & \textbf{Description}   \\ \midrule
\multirow{7}{*}{Ego} & $s$               & m           & Long. coordinate\\
                     & $n$                     & m           & Lat. coordinate \\
                     & $\psi$                  & rad         & Heading angle      \\
                     & $v$                     & m/s         & Velocity         \\
                     & $\kappa$                & 1/m         & Curvature        \\
                     & $a$                     & m/s$^2$       & Acceleration       \\
                     & $\delta$                & 1           & Tire utilization       \\\midrule
\multirow{3}{*}{Opponent} & $s_{\mathrm{rel}}$& m         & Long. position relative to ego            \\
                     & $n_{\mathrm{rel}}$    & m           & Lat. distance relative to ego                 \\
                     & $v_{\mathrm{rel}}$    & m/s         & Velocity relative to ego                      \\ \midrule
\multirow{2}{*}{Track} & $w_{\mathrm{tr, l/r}}(s)$& m       & Lat. distance to boundaries      \\
                     & \textbf{G}($s$)& various     & Reference line 3D geometry \\
\bottomrule
\end{tabular}
\end{table}

\textbf{Action Space:} The action space $\mathcal{A}$ is discrete and represents a library of pre-defined behavioral modes. Each action $a_{t} \in \mathcal{A}$ corresponds to selecting a complete set of cost function weights $W = \{w_{\mathrm{rl}}$, $w_{\mathrm{v}}$, $w_{\mathrm{a}}$, $w_{\mathrm{pr}}$, $w_{\mathrm{c}}$\} for the low-level planner (eq. \eqref{eq:cost_fun}). We define three distinct sets of cost function weights representing different tactical behaviors: 
\begin{itemize}
    \item \textbf{\ac{NR}}, representing a balanced approach, suitable for multi-vehicle racing that prioritizes safety over close overtaking.
    \item \textbf{\ac{AG}}, encouraging proactive overtaking maneuvers by reducing collision-related costs, but increasing the risk of a close encounter.
    \item \textbf{\ac{CD}}, enabling a close drive-by by setting all cost terms except for the collision cost to a low value.
\end{itemize}

\textbf{Reward Design:} The agent's learning process is guided by a reward function that encourages both competitive performance and safe behavior. The total reward $R_t$ at each iteration is a weighted sum of several dense rewards and a sparse reward given at the end of an episode:
\begin{equation}
    R_t = \sum_{i=1}^{\mathcal{D}} w_{i, \text{dense}} \, R_{i, \text{dense}} + w_{\text{sparse}} \, R_{\text{sparse}},
\end{equation}
where $w_{i, \text{dense}}$ and $w_{\text{sparse}}$ are fixed weights, and $\mathcal{D}=5$ is the number of dense reward terms.
Table \ref{tab:reward_design} gives an overview of the used rewards. The dense rewards guide the agent's immediate actions. Among these, the progress reward ($R_{\mathrm{p}}$) incentivizes forward movement. Outside of interactions, the agent is encouraged to follow the optimal racing line by penalizing deviations in velocity ($R_{\mathrm{v}}$) and lateral position ($R_{\mathrm{lat}}$).
To enable proactive overtaking maneuvers, we introduce a dynamic reward structure based on a predefined interaction zone around the opponent vehicle, as highlighted in Figure~\ref{fig:scene_design}. When the ego vehicle enters this zone, the reward logic adapts so that the gap reward ($R_{\mathrm{gap}}$) and collision penalty ($R_{\mathrm{col}}$) become active, focusing the agent on managing its proximity to the opponent while remaining safe. The lateral raceline tracking term ($R_{\mathrm{lat}}$) is temporarily suppressed to allow the agent to dynamically re-plan new overtaking maneuvers.
Finally, a large sparse reward ($R_{\mathrm{lap}}$) is given at the end of an episode to reinforce the overall outcome: a scenario is considered successful if an overtake has been completed and a sufficient longitudinal gap to the opponent is created. A scenario fails if the planner is unable to find a valid trajectory or if the vehicle leaves the track, resulting in a negative reward.

\begin{table}[h!]
\centering
\caption{Formulation of the reward function terms.}
\label{tab:reward_design}
\renewcommand{\arraystretch}{1.4}
\begin{tabular}{@{}ll@{}}
\toprule
\textbf{Reward Term} & \textbf{Formulation} \\
\midrule
\multicolumn{2}{l}{\textit{\textbf{Dense Rewards}}} \\
Progress ($R_{\mathrm{p}}$) & $s_{t} - s_{t-1}$ \\
Velocity Deviation ($R_{\mathrm{v}}$) & $-|v_{\text{target}}(t) - v(t)|$ \\
Lateral Deviation ($R_{\mathrm{lat}}$) & $-(n(t))^2$ \\
Gap to Opponent ($R_{\mathrm{gap}}$) & $-(s_{\text{ego}}(t) - s_{\text{opp}}(t))$ \\
Collision Penalty ($R_{\mathrm{col}}$) & $-D_{\text{lat}}(t)$  \\
\midrule
\multicolumn{2}{l}{\textit{\textbf{Sparse Reward}}} \\
Episode Outcome ($R_{\mathrm{lap}}$) & 
    $\begin{cases} +C & \text{on success} \\ -C & \text{on failure} \end{cases}$ \\
\bottomrule
\end{tabular}
\end{table}

\textbf{\ac{RL}-Agent Architecture:} The high-level behavior selector is a \ac{PPO} \cite{schulman2017Proximal} agent, an on-policy actor-critic algorithm known for its stability and performance. The agent's architecture consists of feature encoders and the policy/value networks. To process the heterogeneous state information, we use separate neural network encoders for the ego, opponent, and track data. The track information is processed by a 1D \ac{CNN} to effectively capture spatial and temporal patterns as shown in Figure~\ref{fig:method_overview}. The outputs of all encoders are concatenated into a single feature vector. The concatenated feature vector serves as input to two fully connected networks: the policy network $\pi_{\theta}(a_t|s_t)$ (actor), which outputs a probability distribution over the discrete action space, and the value network $V_{\phi}(s_t)$ (critic), which estimates the expected return from a given state.

\textbf{Training Objective:} The agent is trained by optimizing the clipped surrogate objective function of the \ac{PPO} algorithm, which is designed to prevent large, destabilizing policy updates. The objective function is given by:
\begin{equation}
L^{\text{CLIP}}(\theta) = \mathbb{E}_{t}\big[\min\big(r_{t}(\theta)\hat{A}_{t}, \text{clip}(r_{t}(\theta), 1-\epsilon, 1+\epsilon)\hat{A}_{t}\big)\big]
\end{equation}
where $r_t(\theta) = \frac{\pi_{\theta}(a_t|s_t)}{\pi_{\theta_{\text{old}}}(a_t|s_t)}$ is the probability ratio between the current and old policies, $\hat{A}_t$ is an estimate of the advantage function at timestep $t$, and $\epsilon$ is a hyperparameter defining the clipping range.
The final loss function that is minimized combines the policy objective with a value function loss $L_t^{\text{VF}}$ and an entropy bonus $S$ to encourage exploration:
\begin{equation}
L(\theta, \phi) = \mathbb{E}_t [ -L_t^{\text{CLIP}}(\theta) + c_1 L_t^{\text{VF}}(\phi) - c_2 S[\pi_\theta](s_t) ]
\end{equation}
where $c_1, c_2$ are coefficients.

\textbf{Training Procedure:} The complete training loop follows the standard \ac{PPO} algorithm, where rollouts are collected and used to update the actor and critic network parameters ($\theta$ and $\phi$) via the Adam optimizer \cite{kingma2017Adam} over several epochs. To ensure the agent learns a robust policy, the training includes both \textit{non-interactive} opponents that follow a fixed racing line and \textit{interactive} opponents controlled by our baseline planner with static weights. This forces the agent to learn not only fundamental overtaking maneuvers, but also more complex responses to a reactive opponent. Our RL agent is trained in parallel across multiple simulation environments to increase sample efficiency.

\section{Results \& Discussion}
\label{sec:results}
We conduct experiments in a deterministic, three-dimensional autonomous racing simulation environment on two different racetracks. We assume perfect trajectory tracking to isolate the planner's decision-making from control uncertainties. We compare our \ac{RL}-based planner to a state-of-the-art sampling-based motion planner \cite{ogretmen2024SamplingBasedb}, where we use the three static weight sets of Section \ref{subsec:rl_procedure} for nominal racing (NR), aggressive (AG), and close driving (CD). Table \ref{tab:parameter_sets} lists the cost function weight values selected for each of the three sets.
\begin{table}[h!]
\centering
\caption{Cost function weight sets for each behavioral mode.}
\label{tab:parameter_sets}
\small
\setlength{\tabcolsep}{5pt}
\begin{tabular}{@{}lrrr@{}}
\toprule
\textbf{Cost Weight}  & \textbf{NR} & \textbf{AG} & \textbf{CD}\\
\midrule
$w_{\mathrm{rl}}$  & 50.0 & 1.0 & 1.0 \\
$w_{\mathrm{v}}$  & 10.0 & 10.0  & 1.0\\
$w_{\mathrm{a}}$  & 500.0 & 200.0 & 1.0 \\
$w_{\mathrm{pr}}$  & \num{1e5} & \num{1e4} & 1.0 \\
$w_{\mathrm{c}}$  & \num{1e8} & 1.0  & 100.0\\
\bottomrule
\end{tabular}
\end{table}

The test scenarios for multi-vehicle performance evaluation involve an ego vehicle and one opponent vehicle. To make overtaking difficult but not impossible, the opponent vehicle uses \SI{90}{\percent} of the ego's acceleration limits, resulting in a marginally slower lap time. As performance metrics, we take the collision rate, maneuver time until the ego vehicle is \SI{15}{\meter} in front of an opponent, and how many overtakes can be completed within the distance of one lap. Multiple overtakes per lap are possible because a new opponent is spawned once the ego vehicle is more than \SI{100}{\meter} in front of the opponent.
All training and evaluation runs are performed on an AMD Ryzen 9 7950X 16-Core CPU. For the training, which is performed on Yas Marina Circuit, we use the \ac{PPO} implementation from stable-baselines3 \cite{stable_baselines}. To eliminate the influence of varying runtime on the results, we fix the simulation time step per planning step to \SI{0.35}{\second}.


\begin{figure*}[!t]
   \centering
    \includegraphics[width=0.95\textwidth]{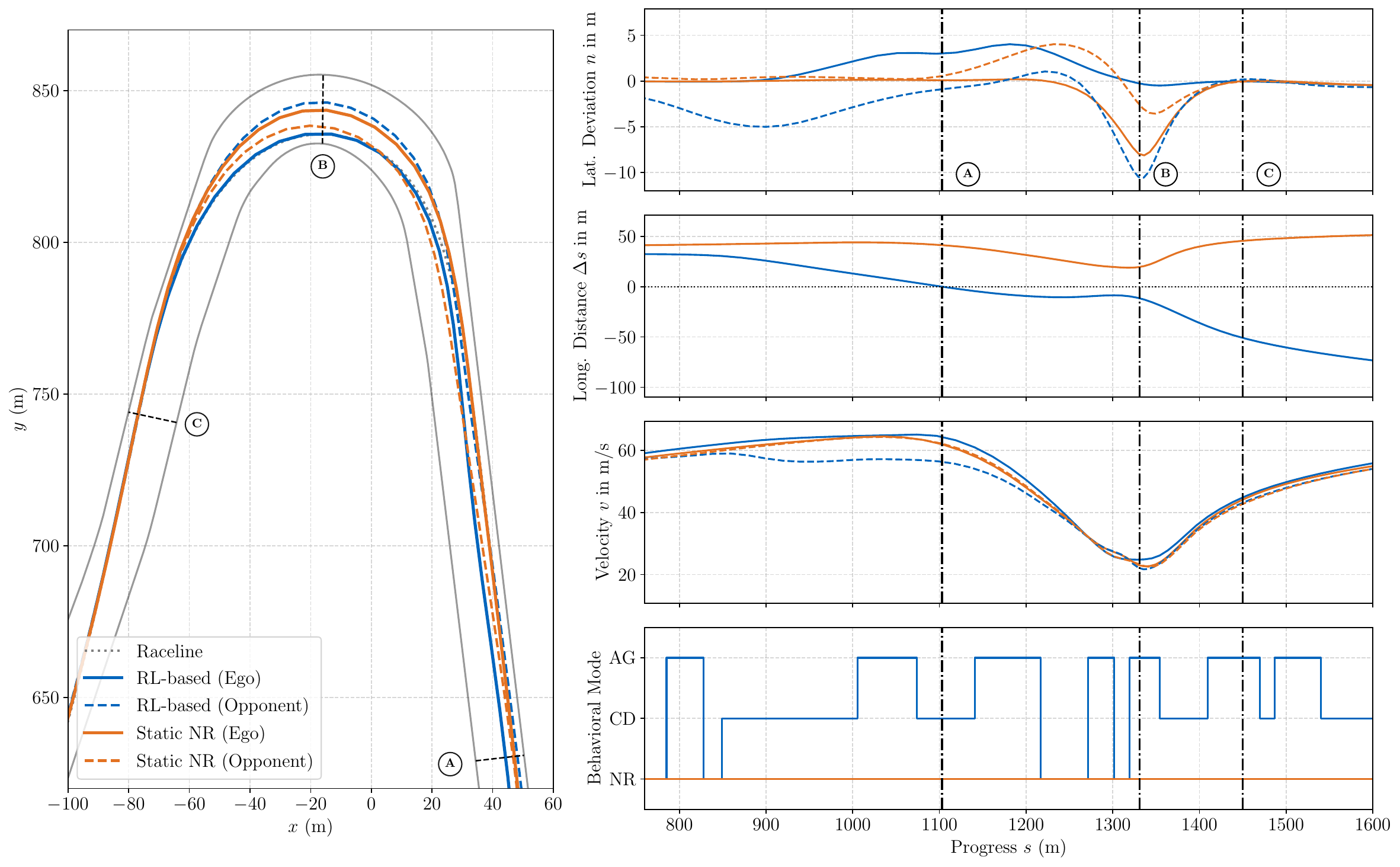}
    \caption{Qualitative analysis of an overtaking maneuver, comparing our RL-based planner against a static-weight planner with the NR parameter set. In both cases, the opponent is a reactive NR-planner. Left: driven trajectories in a section of the analyzed scenario. Right: lateral positions (top), longitudinal gap $\Delta s$ between ego and opponent (upper middle, where $\Delta s>0$ means the opponent is ahead of the ego), velocity profiles (lower middle), and selected parameter set from our planner (bottom) during the scenario.}
    \label{fig:qualitative_analysis}
\end{figure*}

\begin{figure}[!b]
    \centering
    \includegraphics[width=1.0\linewidth]{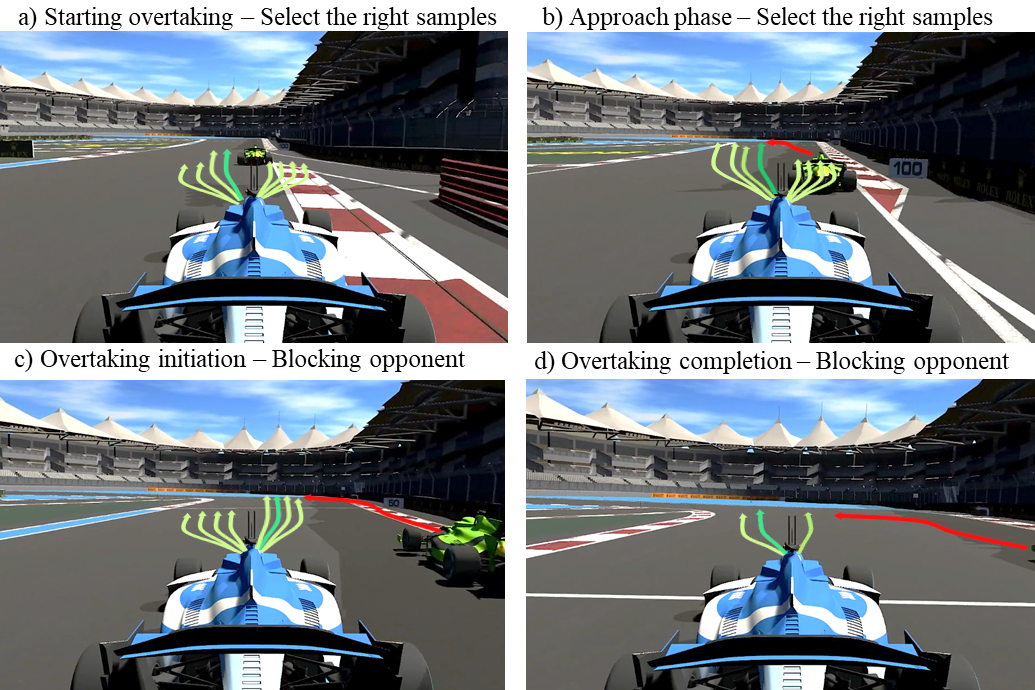}
    \caption{Overtake scenario similar to Figure \ref{fig:qualitative_analysis}, in our 3D simulation environment. Dark green trajectories denote the selected trajectory in each step from the set of available trajectories (light green), and the red trajectory visualizes the opponent's behavior. Note that only a subset of the generated trajectories is shown.}
    \label{fig:3d_sim_overtake}
\end{figure}

\subsection{Quantitative Analysis}
\label{sec:quantitative_results}
We evaluate our planner's performance over 260 distinct scenarios on Yas Marina Circuit with different initial starting positions of the vehicles. For this, we introduce two different opponent types: non-reactive opponents that simply follow their raceline, and reactive opponents that use our baseline planner with the static \ac{NR} cost function. The results of these extensive test campaigns are summarized in Table~\ref{tab:multivehicle_results}.

Against the non-reactive opponent, the results reveal a clear trade-off for the static planners. The \ac{NR} baseline is the safest (\SI{0.0}{\percent} collision rate) but also the least competitive (\SI{20.7}{\second} per successful overtake). Conversely, the \ac{AG} and \ac{CD} baselines are much faster but suffer from high collision rates of \SI{14.4}{\percent} and \SI{29.1}{\percent}, respectively. Our \ac{RL}-based planner resolves this trade-off, achieving \SI{0.0}{\percent} collision rate while being nearly as fast as the most aggressive static planners, with an average maneuver time of \SI{12.0}{\second}.

This advantage becomes even more pronounced against the reactive opponent. While the static \ac{NR} baseline remains safe, it becomes exceedingly cautious, taking an average of \SI{33.3}{\second} per overtake. The aggressive static planner again proves to be unsafe, with a \SI{3}{\percent} collision rate. In contrast, our \ac{RL}-based planner maintains safety with a \SI{0}{\percent} collision rate, and the time until an overtake is completed is \SI{60}{\percent} faster than the NR baseline.
This confirms that the agent learns to leverage the full spectrum of available behaviors, achieving a combination of safety and performance that is unattainable with any single, static weight strategy.

\begin{table}[h!]
    \centering
    \caption{Comparing the collision rate, average overtaking time, and number of overtakes per lap, for our RL-based planner and the static-weight baselines.
    }
    \label{tab:multivehicle_results}
    \small 
    \setlength{\tabcolsep}{3pt} 
    \begin{tabular}{@{}lccc@{}}
        \toprule
        \textbf{Configuration} & \textbf{Collision [\%]} & \textbf{Time [s]} & \textbf{Overtakes/Lap} \\
        \midrule
        \multicolumn{4}{c}{\textit{Opponent: Non-reactive Raceline Follower}} \\
        \midrule
        Nominal Racing \cite{ogretmen2024SamplingBasedb}   & 0.0  & 20.7 & 1.6 \\
        Aggressive \cite{ogretmen2024SamplingBasedb} & 14.4 & 11.2 & 2.7 \\
        Close Driving \cite{ogretmen2024SamplingBasedb}  & 29.1 & 11.8 & 2.6 \\
        \textbf{\ac{RL}-based (ours)}  & \textbf{0.0} & \textbf{12.0} & \textbf{2.2} \\
        \midrule
        \multicolumn{4}{c}{\textit{Opponent: Reactive NR Planner}} \\
        \midrule
        Nominal Racing \cite{ogretmen2024SamplingBasedb}   & 0.0  & 33.3 & 0.6 \\
        Aggressive \cite{ogretmen2024SamplingBasedb} & 3.0 & 19.8 & 1.0 \\
        Close Driving \cite{ogretmen2024SamplingBasedb}  & 18.5 & 13.6 & 1.7 \\
        \textbf{\ac{RL}-based (ours)}  & \textbf{0.0} & \textbf{12.0} & \textbf{1.5} \\
        \bottomrule
    \end{tabular}
\end{table}

\subsection{Qualitative Analysis}
\label{sec:qualitative_analysis}
To provide a deeper insight into the performance difference between our adaptive planner and the static baselines, we conduct a qualitative analysis of a racing scenario against a reactive NR-planner opponent on the Yas Marina circuit. The scenario is shown in Figure~\ref{fig:qualitative_analysis}, and Figure \ref{fig:3d_sim_overtake} depicts qualitatively the scenario in our 3D simulation environment. The ego vehicle is initialized at $s=$ \SI{0}{\meter}, which corresponds to the start-finish line, and the opponent at $s=$ \SI{60}{\meter}. We perform a direct comparison between our \ac{RL}-based planner and the static-weight \ac{NR} baseline.

The scenario begins with both vehicles (our planner and the baseline) approaching the opponent in a corner of the track. By building a lateral offset to the opponent, the RL-based planner forces the opponent to reduce its velocity at $s=$ \SI{850}{\meter} to avoid a collision. This allows the RL-based planner to maintain a higher speed for longer and pass the opponent at Point A. 
Our RL agent temporarily selects the aggressive (\ac{AG}) mode at the beginning of the maneuver (bottom-right plot in Figure \ref{fig:qualitative_analysis}): this lowers the planner's penalty for proximity and collision risk, allowing it to force the reactive opponent to a yielding maneuver. This enables planning to the faster inside line of the turn (Point B). It then completes the overtake by returning to the raceline (Point C). 
In contrast, the overtake cannot be achieved by the NR-baseline, as it expects a collision and cannot anticipate the yielding maneuver by the opponent when committing to an aggressive trajectory. As a result, it favors staying behind the opponent.

This qualitative example provides evidence that the agent's learned ability to dynamically adapt its cost function parameters and, implicitly, its risk profile, is a key factor in its improved interactive performance.

\subsection{Generalization to an Unseen Racetrack}
To assess the generalization capabilities of the learned policy, we evaluate its performance on an unseen racetrack, the Autodromo Nazionale Monza, without any retraining. We test the agent in the same scenario setup as in Section \ref{sec:quantitative_results} against a reactive opponent controlled by the \ac{NR}-planner. The results, presented in Table~\ref{tab:monza_results}, show that the agent successfully transfers its learned behavioral skills.

\begin{table}[b!]
    \centering
    \caption{Performance on the unseen Monza track against a reactive opponent.}
    \label{tab:monza_results}
    \small 
    \setlength{\tabcolsep}{3pt} 
    \begin{tabular}{@{}lccc@{}}
        \toprule
        \textbf{Configuration} & \textbf{Collision [\%]} & \textbf{Time [s]} & \textbf{Overtakes/Lap} \\
        \midrule
        Nominal Racing \cite{ogretmen2024SamplingBasedb}   & 0.0  & 22.0 & 0.4 \\
        Aggressive \cite{ogretmen2024SamplingBasedb} & 3.1 & 15.9 & 0.8 \\
        Close Driving \cite{ogretmen2024SamplingBasedb}  & 24.0 & 11.0 & 1.3 \\
        \textbf{\ac{RL}-based (ours)}  & \textbf{0.0} & \textbf{15.9} & \textbf{1.3} \\
        \bottomrule
    \end{tabular}
\end{table}

While all static planners exhibit a degradation in either safety or performance on the new track, our \ac{RL}-based planner maintains a \SI{0.0}{\percent} collision rate while performing the most overtakes among all tested planners. This indicates that the agent is able to generalize and does not overfit to the layout of the track that was used in the training process.

\subsection{Single-vehicle Performance Evaluation}
We evaluate all configurations in a single-vehicle scenario to establish a performance baseline on a clear track. The results are reported in Table~\ref{tab:lap_times}: the parameters of the static-weight \ac{CD} configuration are most favorable for this task as it achieves the fastest lap time. Our \ac{RL}-based planner demonstrates competitive performance, with a lap time only \SI{0.35}{\second} slower.

\begin{table}[ht]
    \centering
    \caption{Single-vehicle Lap Times.}
    \label{tab:lap_times}
    \begin{tabular}{@{}lc@{}}
        \toprule
        \textbf{Planner Configuration} & \textbf{Lap Time [s]} \\
        \midrule
        Nominal Racing (NR) & \SI{112.35}{} \\
        Aggressive (AG) & \SI{113.40}{} \\
        Close Driving (CD) & \SI{111.65}{} \\
        \textbf{\ac{RL}-based (ours)} & \textbf{\SI{112.00}{}} \\
        \bottomrule
    \end{tabular}
\end{table}

This outcome is expected, as the single-vehicle scenario represents a deterministic problem that lacks the complex and uncertain behavior introduced by other vehicles. In such a predictable environment, a static parameter set can be manually optimized for the singular objective of minimizing lap time, yielding near-optimal results. In contrast, the policy for our \ac{RL}-based planner was trained across a wider distribution of scenarios, including challenging multi-vehicle interactions. 
The minimal performance loss demonstrates that the behavioral adaptability developed for complex, interactive situations does not introduce a significant performance penalty in the simpler, static case. The policy predominantly selects the appropriate high-performance behavioral modes (\ac{CD} or \ac{NR}).

\subsection{Runtime Evaluation}
We analyze the computational cost of our approach, focusing on the additional overhead introduced by the RL agent's inference step. The underlying sampling-based planner, which forms the basis for all configurations, has a mean computation time of \SI{102}{\milli\second} per cycle. Our measurements show that the RL agent's policy inference adds a minimal overhead, with an average prediction time of only \SI{3.031}{\milli\second} and a minimum of \SI{1.895}{\milli\second}. This shows that the significant benefits in tactical adaptability and interactive capability are achieved with a negligible impact on the system's overall computational load.

\subsection{Limitations}
While our results demonstrate a clear performance benefit, we identify three main limitations to our current study. First, our evaluation is conducted against opponents with predictable behavior (either non-reactive or following a fixed, reactive policy). While this setup enabled consistent benchmarking, it does not capture the variability of fully strategic agents. For example, against our reactive NR baseline, the RL-based planner achieved an average overtake time of \SI{12.0}{\second}, whereas the NR baseline required \SI{33.3}{\second}. However, these gains may diminish or change qualitatively when facing adaptive real-world opponents. The framework's performance against fully strategic, game-theoretic adversaries, such as other learning-based agents, remains an open question for future work.  

Second, the agent's learned policy represents a compromise across the diverse scenarios encountered during training. As shown in the single-vehicle evaluation, this can lead to a marginally slower lap time compared to a static parameter set. The agent's parameter selection is therefore suboptimal for this case because its policy is generalized to handle more complex, interactive situations that are not present on a clear track.

Finally, all experiments were conducted in a deterministic simulation with perfect state information and trajectory tracking. The robustness of the learned policy to real-world challenges, such as sensor noise, state estimation errors, and vehicle dynamics uncertainties, must be validated through future hardware deployments.

\section{Conclusion \& Future Work}
In this work, we addressed a fundamental limitation of classical sampling-based motion planners in autonomous racing: their reliance on a static set of cost function parameters, which leads to purely reactive and tactically inflexible behavior. We argued that this forces a suboptimal compromise, preventing the vehicle from executing the proactive, interactive maneuvers essential for competitive performance.

To overcome this, we presented a novel hybrid planning architecture that integrates a high-level \ac{RL} agent with a low-level sampling-based trajectory planner. The RL agent learns to function as a behavior selector, dynamically switching between a predefined library of behavior-specific cost function parameter sets depending on the race context. 
Our experiments on Yas Marina Circuit demonstrated that this dynamic, learned adaptation significantly outperforms the baseline planner operating with a single, static set of weights in interactive multi-vehicle scenarios. The results confirm that by separating high-level decision-making from low-level trajectory generation, our hybrid system achieves a level of performance and interactivity unattainable by its classical counterpart alone. We also showed the generalization capabilities of our approach by applying the agent in Monza, which was not included in the training process.

A next step for future work is to validate this framework on a full-scale autonomous race car, bridging the gap from simulation to real-world application. Further research will also focus on extending this framework to scenarios with a higher number of opponents that also act interactively by the use of, e.g., game theory. This research provides a clear methodology for creating more adaptive yet interpretable and trustworthy motion planning systems for safety-critical autonomous applications.


\bibliographystyle{IEEEtran}
\bibliography{literature.bib}

\begin{acronym}
\acro{AVs}{autonomous vehicles}
\acro{RL}{reinforcement learning}
\acro{PPO}{proximal policy optimization}
\acro{CNN}{convolutional neural network}
\acro{CD}{Close Driving}
\acro{NR}{Nominal Racing}
\acro{AG}{Aggressive}
\acro{MDP}{Markov Decision Process}
\end{acronym}


\end{document}